\documentclass{article}

\usepackage[numbers,sort, compress]{natbib}
\usepackage[preprint]{neurips_2020}

\usepackage[utf8]{inputenc} 
\usepackage[T1]{fontenc}    
\usepackage{hyperref}       
\usepackage{url}            
\usepackage{booktabs}       
\usepackage{amsfonts}       
\usepackage{nicefrac}       
\usepackage{microtype}      

\usepackage{cite}
\usepackage{amsmath,amssymb,amsfonts}
\usepackage{algorithm}
\usepackage{graphicx}
\usepackage{textcomp}
\usepackage{xcolor}
\usepackage{times}
\usepackage{epsfig}
\usepackage{graphicx}
\usepackage{amsmath}
\usepackage{amssymb}
\usepackage{algorithm}
\usepackage{algorithmicx}
\usepackage{microtype}
\usepackage[noend]{algpseudocode}

\usepackage{fancyhdr}

\title{Mixture of Step Returns in Bootstrapped DQN}

\author{%
  David S.~Hippocampus\thanks{Use footnote for providing further information
    about author (webpage, alternative address)---\emph{not} for acknowledging
    funding agencies.} \\
  Department of Computer Science\\
  Cranberry-Lemon University\\
  Pittsburgh, PA 15213 \\
  \texttt{hippo@cs.cranberry-lemon.edu} \\
}

\makeatletter
\newcommand{\printfnsymbol}[1]{%
  \textsuperscript{\@fnsymbol{#1}}%
}
\makeatother
\author{%
Po-Han Chiang\thanks{Equal contribution}
\And
Hsuan-Kung Yang\printfnsymbol{1}\And
Zhang-Wei Hong\And
Chun-Yi Lee\AND
\normalfont{Elsa Lab, Department of Computer Science}\\
National Tsing Hua University\\
Hsinchu, Taiwan\\
\texttt{\{ymmoy999, hellochick, williamd4112, cylee\}@gapp.nthu.edu.tw}
}

\begin{document}

\maketitle

\thispagestyle{fancy} 
\pagestyle{fancy}
\fancyhf{} 
\renewcommand{\headrulewidth}{0pt}
\fancyfoot[C]{\thepage}

\begin{abstract}

The concept of utilizing multi-step returns for updating value functions has been adopted in deep reinforcement learning (DRL) for a number of years. Updating value functions with different backup lengths provides advantages in different aspects, including bias and variance of value estimates, convergence speed, and exploration behavior of the agent. Conventional methods such as TD~($\lambda$) leverage these advantages by using a target value equivalent to an exponential average of different step returns. Nevertheless, integrating step returns into a single target sacrifices the diversity of the advantages offered by different step return targets. To address this issue, we propose Mixture Bootstrapped DQN (MB-DQN) built on top of bootstrapped DQN, and uses different backup lengths for different bootstrapped heads. MB-DQN enables heterogeneity of the target values that is unavailable in approaches relying only on a single target value. As a result, it is able to maintain the advantages offered by different backup lengths. In this paper, we first discuss the motivational insights through a simple maze environment. In order to validate the effectiveness of MB-DQN, we perform experiments on the \textit{Atari 2600} benchmark environments, and demonstrate the performance improvement of MB-DQN over a number of baseline methods. We further provide a set of ablation studies to examine the impacts of different design configurations of MB-DQN.

\end{abstract}

\section{Introduction}
\label{sec::introduction}





In recent value-based deep reinforcement learning (DRL), a value function is usually utilized to evaluate state values, which stands for estimates of the expected long-term cumulative rewards that might be collected by an agent. In order to perform such an evaluation, a deep neural network (DNN) is employed by a number of contemporary value-based DRL methods \citep{dqn,duelingDQN,doubleDQN,BstrapDQN,rainbowDQN} as the value function approximator, in which the network parameters are iteratively updated based on the agent's experience of interactions with an environment. For many of these methods~\citep{dqn,duelingDQN,doubleDQN,BstrapDQN,rainbowDQN}, the update procedure is carried out by one-step temporal-difference (TD) learning~\citep{Sutton1998} (or simply ``\textit{one-step TD}'' hereafter), which calculates the error between an estimated state value and a target differing by one timestep. One-step TD has been demonstrated effective in backing up immediate reward signals collected by an agent. Nevertheless, the long temporal horizon that the reward signals from farther states have to propagate through might lead to an extended learning period of the value function approximator.

Learning from multi-step returns~\citep{Sutton1998} is a way of propagating rewards newly observed by the agent faster to earlier visited states, and has been adopted in several previous works. Asynchronous advantage actor-critic (A3C)~\citep{a3c} employs multi-step returns as targets to update the value functions of its asynchronous threads. Rainbow deep Q-network (Rainbow DQN)~\citep{rainbowDQN} also utilizes multi-step returns during the backup procedure. The authors in~\citep{D4PG} also modify the target value function of deep deterministic dolicy gradient (DDPG)~\citep{DDPG} to estimate TD errors using multi-step returns. Updating value functions with different backup lengths provides advantages in different aspects, including bias and variance of value estimates, convergence speed, and exploration behavior of the agent. Backing up reward signals through multi-step returns shifts the bias-variance tradeoff~\citep{rainbowDQN}. Therefore, backing up with different step return lengths (or simply `\textit{backup length}' hereafter~\citep{unifying}) might lead to different target values in the Bellman equation, resulting in different exploration behaviors of the agent as well as different achievable performance of it. The authors in~\citep{TDnotTD} have demonstrated that the performance of the agent varies with different backup lengths, and showed that both very short and very long backup lengths could cause performance drops. These insights suggest that identifying the best backup length for an environment is not straightforward. 
In addition, although learning based on multi-step returns enhances the immediate sensitivity to future rewards, it is at the expense of greater variance which may cause the value function approximator to require more data samples to converge to the true expectation. Moreover, relying on a single target value with any specific backup length constrains the exploration behaviors of the agent, and might limit the achievable performance of it.




Based on the above observations, there have been several research works proposed to unify different target values with different backup lengths to leverages their respective advantages.  The traditional TD~($\lambda$)~\citep{Sutton1998} uses a target value equivalent to an exponential average of all $n$-step returns (where $n$ is a natural number), providing a faster empirical convergence by interpolating between low-variance TD returns and low-bias Monte Carlo returns. DQN~($\lambda$)~\citep{DQNlambda} further proposes an efficient implementation of TD~($\lambda$) for DRL by modifying the replay buffer memory such that $\lambda$-returns can be pre-computed. Although these methods benefit from combining multiple distinct backup lengths, they still rely on a single target value during the update procedure. Integrating step returns into a single target value, nevertheless, may sacrifice the diversity of the advantages provided by different step return targets.

As a result, in this paper we propose \textbf{M}ixture \textbf{B}ootstrapped \textbf{DQN} (abbreviated as ``\textit{MB-DQN}''), to address the above issues. MB-DQN is built on top of bootstrapped DQN~\citep{BstrapDQN}, which contains multiple bootstrapped heads with randomly initialized weights to learn a set of value functions. MB-DQN leverages the advantages of different step return targets by assigning a distinct backup length to each bootstrapped head. Each bootstrapped head maintains its own target value derived from the assigned backup length during the update procedure. Since the backup lengths of the bootstrapped heads are distinct from each other, MB-DQN provides heterogeneity in the target values as well as diversified exploration behaviors of the agent that is unavailable in approaches relying only on a single target value. To validate the proposed concept, in our experiments, we first provide motivational insights on the influence of different configurations of backup lengths in a simple maze environment. We then evaluate the proposed MB-DQN on the \textit{Atari 2600}~\citep{Atari} benchmark environments, and demonstrate its performance improvement over a number of baseline methods. We further provide a set of ablation studies to analyze the impacts of different design configurations of MB-DQN. In summary, the primary contributions of this paper include: (1) introducing an approach for maintaining the advantages from different backup lengths, (2) providing heterogeneity in the target values by utilizing multiple bootstrapped heads, and (3) enabling diversified exploration behaviors of the agent. 

The remainder of this paper is organized as the following. Section~\ref{sec::background} provides the background material related to this work. Section~\ref{sec::methodology} walks through the proposed MB-DQN methodology. Section~\ref{sec::experiments} reports the experimental results, and presents a set of the ablation analyses. Section~\ref{sec::conclusion} concludes this paper.
\section{Background}
\label{sec::background}

In this section, we provide the background material related to this work. We first introduce the basic concepts of the Markov Decision Process~(MDP) and one-step return, followed by an explanation of the concept of multi-step returns. Next, we provide a brief overview of the Deep Q-Network (DQN).

\subsection{Markov Decision Process and One-Step Return}
\label{subsec::MDP_One_Step_Return}

In RL, an agent interacting with an environment $\mathcal{E}$ with state space $\mathcal{S}$ and action space $\mathcal{A}$ is often formulated as an MDP. At each timestep $t$, the agent perceives a state \textit{$s_t$} $\mathcal{\in S}$, takes an action $a_t$ $\mathcal{\in A}$ according to its policy $\pi(a|s)$, receives a reward $r_t \sim R(s_t, a_t)$, and transits to next state $s_{t+1} \sim p(s_{t+1}|s_t, a_t)$, where $R(s_t, a_t)$ and $p(s_{t+1}|s_t, a_t)$ are the reward function and transition probability function, respectively. The main objective of the agent is to learn an optimal policy $\pi^*(a|s)$ that maximizes discounted cumulative return $G_{t} = \sum_{i=t}^{T} \gamma^{i-t}r_t$, 
where $\gamma \in (0, 1]$ is the discount factor and $T$ is the horizon. For a given policy $\pi(a|s)$, the state value function $V^{\pi}$ and state-action value function $Q^{\pi}$ are defined as the expected discounted cumulative return $G_{t}$ starting from a state $s$ and a state-action pair~$(s, a)$ respectively, and can be represented as the following:
\begin{equation}
\label{eqa::}
 V^{\pi}(s) = \mathbb{E}[G_t|s_t = s, \pi], Q^{\pi}(s, a) = \mathbb{E}[G_t|s_t = s, a_t = a, \pi].
\end{equation}
In order to maximize $\mathbb{E}[G_t]$, conventional value-based RL methods often use one-step TD learning to iteratively update $V^{\pi}$ and $Q^{\pi}$. Take $Q^{\pi}$ for example, the update rule is expressed as the following:
\begin{equation}
\label{equation1}
    Q(s_t, a_t) \leftarrow Q(s_t, a_t) + \alpha[r_t + \gamma Q(s_{t+1}, a_{t+1}) -Q(s_t, a_t)],
\end{equation}
where $\alpha \in (0, 1]$ is a step size parameter which controls the update speed. This update procedure only considers the immediate return $r_t$ and $\gamma Q(s_{t+1}, a_{t+1})$, which is together called one-step return. 





\subsection{Multi-Step Return}
\label{subsec::Multi_Step_Return}

Multi-step return is a variant of one-step return presented in the previous section. Multi-step return modifies the target of one-step return through bootstrapping over longer time intervals. It replaces the single reward $r_t$ in Eq.~(\ref{equation1}) with the truncated multi-step return $R_{t}^{n}$, which is represented as follows:
\begin{equation}
    R_{t}^{n} = \sum_{j=0}^{n}\gamma^{j}r_{t+j},
\label{eq::multi-step_return}
\end{equation}
where $n$ is the selected backup length. The update rule of Eq.~(\ref{equation1}) is then re-written as the following:
\begin{equation}
    Q(s_t, a_t) \leftarrow Q(s_t, a_t) + \alpha[R_{t}^{n} + \gamma^{n+1} Q(s_{t+n+1}, a_{t+n+1}) -Q(s_t, a_t)].
\end{equation}
A longer backup length $n$ has been shown to increase the variance of the estimated $Q(s_t, a_t)$ as well as decrease its bias~\citep{VarianceTradeoff}. Despite of the high-variance and the increased computational cost, multi-step return enhances the immediate sensitivity of the value approximator to future rewards, and allows them to backup faster. As a result, in certain cases, it is possible to achieve a faster learning speed for the value approximator by using an appropriate backup length $n$ larger than one~\citep{Sutton1998, rainbowDQN, TDnotTD}.

\subsection{Deep Q-Network}
\label{subsec::Deep_Q_Network}

DQN is a DNN parameterized by $\theta$ for approximating the optimal Q-function. DQN is trained using samples drawn from an experience replay buffer $Z$, and is updated based on one-step TD learning with an objective to minimize a loss function $L_{DQN}$, which is typically expressed as the following:
\begin{equation}
    L_{DQN} = \mathbb{E}_{s,a,r,s^\prime \sim U(Z)}\big[(y_{s,a} - Q(s,a, \theta))^2\big],
    \label{eq::dqnloss}
\end{equation}
where $y_{s,a} = r_t + \gamma\max_{a} Q(s_{t+1},a, \theta^{-})$ is the one-step target value, $U(Z)$ is a uniform distribution over $Z$, and $\theta^{-}$ is the parameters of the target network. $\theta^{-}$ is updated by $\theta$ at predefined intervals.




\begin{figure}[b]
    \centering
    \includegraphics[width=\linewidth]{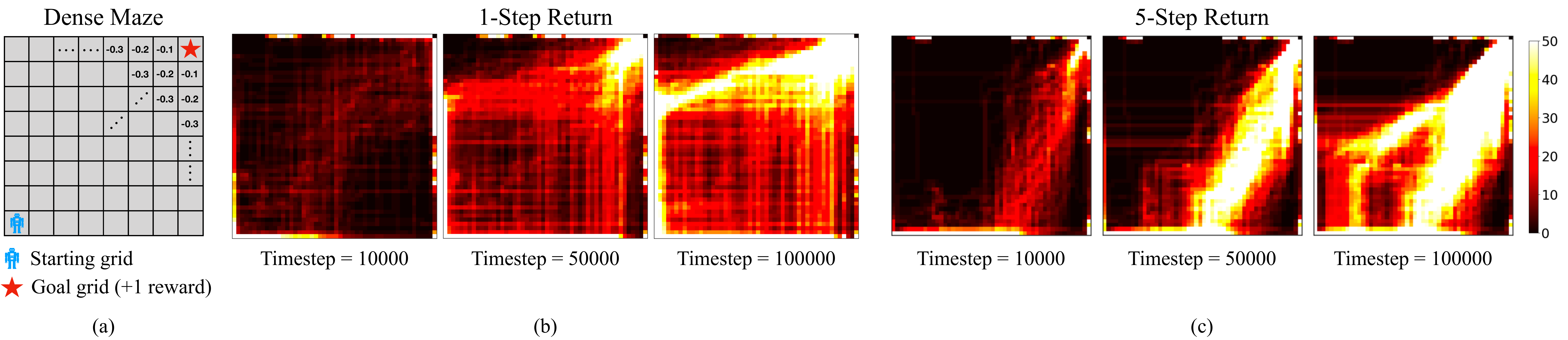}
    \caption{A visualization of the behaviors of the agents with different backup lengths. (a) presents the layout of the maze environment (denoted as \textit{Dense Maze}), which contains a starting grid and a goal grid. (b) and (c) illustrate the behaviors of the agents updated using 1-step return and 5-step return, respectively. It is observed that the agent trained with 5-step return reaches the goal through shorter diagonal trajectories, while the agent trained with 1-step return explores more grids in the early stage.}
    \label{fig:motivation}
\end{figure}

\section{Methodology}
\label{sec::methodology}
In this section, we first demonstrate the impacts of different backup lengths on the behaviors of an agent in a simple maze environment. Then, we walk through the details of the MB-DQN framework.

\subsection{Agent Behavior with Different Backup Lengths in DQN}
\label{subsec:agent_behavior}



To illustrate the impacts of different backup lengths on an agent's behavior, we first consider a toy model in a two-dimensional maze environment containing a starting point and a goal, as depicted in Fig.~\ref{fig:motivation} (a). We use DQN as our default agent and perform our experiments on this maze environment (denoted as \textit{Dense Maze}) with dense rewards. In this setting, the reward of a grid gradually decreases as its distance to the goal increases. We depict the states visited by the agents for 100k timesteps in the training phase in Figs.~\ref{fig:motivation}~(b) and~(c). It is observed that the agent trained with 5-step return reaches the goal through a shorter path than that of the 1-step return case. This is because the longer backup length allows the agent to adjust its value function estimation faster. On the other hand, although the agent trained with 1-step return might converge slower than that of the 5-step return case, it is observed that 1-step return enables the agent to visit and explore more states in the early stage. This is because the reward signal from a farther state has to propagate through a longer temporal horizon. 
Therefore, the agent explores more extensively before learning an effective policy to reach the goal.

\begin{figure}[h]
    \centering
    \includegraphics[width=\linewidth]{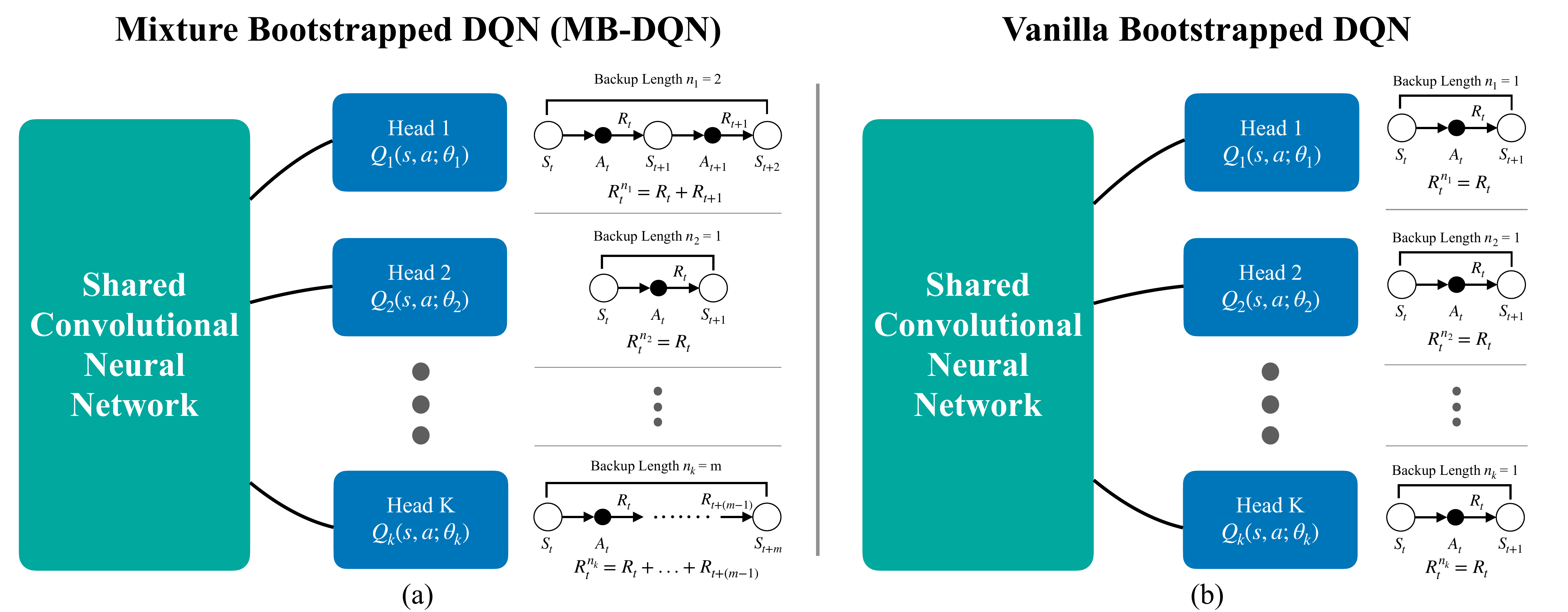}
    \caption{Overview of the proposed MB-DQN framework.}
    \label{fig:archi}
\end{figure}

\subsection{Mixture of Step Returns in Bootstrapped DQN}
\label{subsec:MBDQN}
%
%

In order to combine step returns with different backup lengths, we choose bootstrapped DQN~\citep{BstrapDQN} as our backbone framework. Bootstrapped DQN modifies DQN to approximate a distribution over Q-values via bootstrapping, and has demonstrated both the improved learning speed as well as the performance of the agents in various environments. At the beginning of an episode, bootstrapped DQN uniformly samples a Q-value function head $Q_k(s, a; \theta_k)$, $k \in \{1, ... ,K\}$ from its $K$ bootstrapped Q-value function heads, as shown in Fig.~\ref{fig:archi}. The agent then performs its control according to $Q_k(s, a; \theta_k)$ during the entire episode. Bootstrapped DQN re-samples a Q-value function head for each episode based on the same backup length (i.e., 1-step return) to calculate the target value (i.e., Eq.~(\ref{equation1})). The framework, nevertheless, might be lack in diversity and heterogeneity among the bootstrapped heads.  



As a result, in this paper, we leverage the advantages of distinct Q-value function heads in bootstrapped DQN, and propose the usage of mixture backup lengths for different bootstrapped Q-value function heads in our MB-DQN framework, which is shown in Fig.~\ref{fig:archi}. MB-DQN is similarly implemented as $K$ bootstrapped heads for estimating the Q-value function, where each bootstrapped head $k\in K$ correspond to its own backup lengths $n_k$. In each episode, MB-DQN also uniformly and randomly selects a head $k \in \{1, ... ,K\}$, and stores the state transition data collected by the agent using this head into a replay buffer. The replay buffer is played back periodically to update the parameters of all the bootstrapped Q-value function heads as well as the shared convolutional neural network. Each head is trained with its own target network $Q_k(s, a; \theta^-_k)$ and its own target value $y^k_{s,a}$ with the truncated multi-step return defined in Eq.~(\ref{eq::multi-step_return}). The detailed update methodology is summarized in Algorithm~\ref{Algo::Target_Update_MB-DQN}, while the training methodology is the same as bootstrapped DQN~\citep{BstrapDQN}. The truncated multi-step returns with different backup lengths thus provide diversity and heterogeneity for the $K$ bootstrapped estimates, which balance the strengths and the weaknesses of different backup lengths. 



\begin{algorithm}[h]
\caption{Update Methodology of MB-DQN}
\label{Algo::Target_Update_MB-DQN}
\footnotesize
\begin{algorithmic}[1]
    \State Initialize K Q-networks $Q_{k}(s, a; \theta_k)$ with random weights $\theta_k$
    \State Let each networks $Q_{k}$ with its own backup length $n_{k}$
    \For{each update time}
    \For{each head k = 1, 2,...,K}
        \State $R_{t}^{n_k}$ = $\sum_{j=0}^{n_k}\gamma^{j}R_{t+j}$
        \State $y^{k}_{s,a}$ = $R_{t}^{n_k} + \gamma^{n_k} \max_{a}{Q_{k}{(s_{t+n_{k}},{\arg\max}_{a}{Q(s_{t+n_{k}},a; \theta_k)}; \theta^-_k)}}$
        \State $\theta_k \approx \mathop{\arg\min}_{\theta_k} \mathbf{E}(y^{k}_{s,a} - Q_{k}{(s,a; \theta_k)})$
    \EndFor
    \EndFor
\end{algorithmic}
\end{algorithm}




\section{Experimental Results}
\label{sec::experiments}

In this section, we present the experimental results to demonstrate the advantages of the mixture usage of different backup lengths. We first evaluate the proposed MB-DQN on a collection of well-known \textit{Atari 2600}~\citep{Atari} games, and compare its performance to different configurations of boostrapped DQN both quantitatively and qualitatively in Section~\ref{subsec::atari}. Next, we investigate the quality of the data samples collected by MB-DQN, and demonstrate their advantages in training an RL agent in Section~\ref{subsec::data_sample_quality}. Then, we validate the assumption made in Section~\ref{sec::introduction} that unifying different step return targets to a single target value may not be as effective as the proposed MB-DQN approach  in Section~\ref{subsec::target_comparison}. Finally, we further provide a set of ablation analyses of MB-DQN on \textit{Atari} games to inspect and discuss the impacts of different configurations on MB-DQN's performance in Section~\ref{subsec::ablation}.



\addtocounter{footnote}{-2} 
\begin{figure}[h]
    \centering
    \includegraphics[width=.95\linewidth]{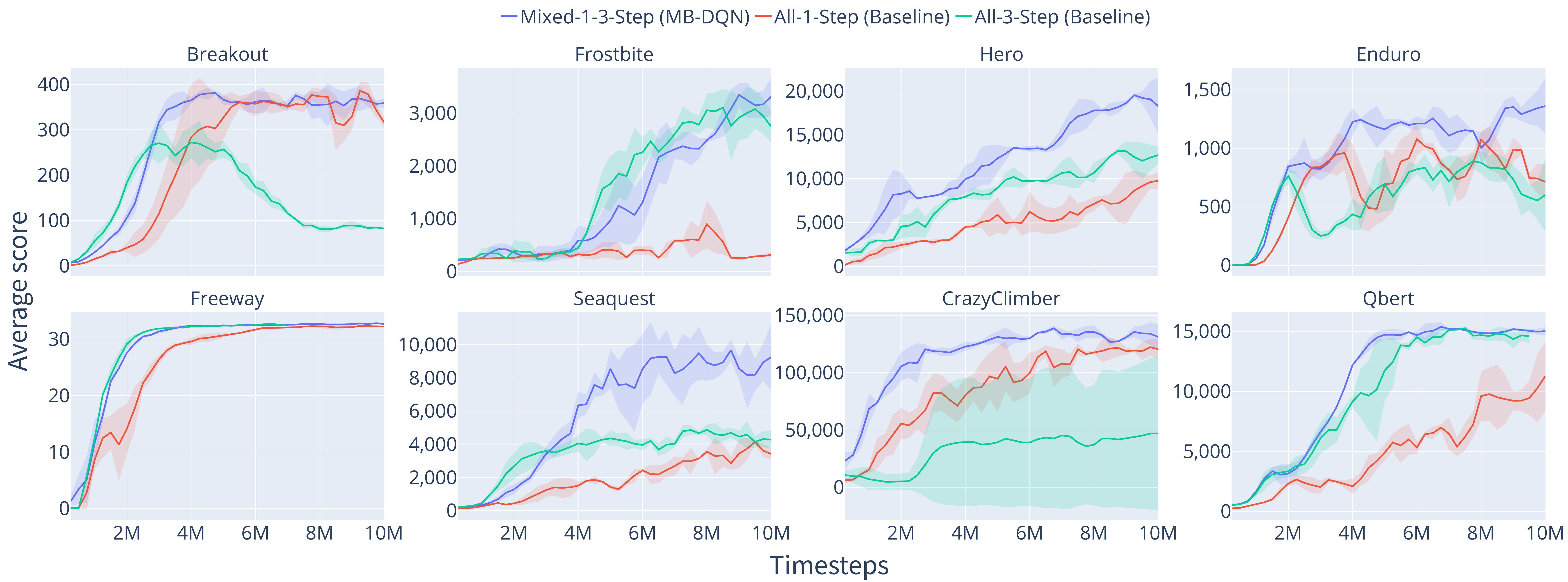}
    \caption{Comparison of the evaluation curves of MB-DQN and the baselines in eight \textit{Atari} games.~\protect\footnotemark}
    \label{fig:atari}
\end{figure}

\footnotetext{For all our experiments, the MB-DQN and the bootstrapped DQN agents are evaluated every 250k timesteps based on the results voted by the majority of their bootstrapped heads. The evaluation curves are averaged from three random seeds and drawn with 68\% confidence interval as the shaded areas.} 

\subsection{Comparison of MB-DQN and Bootstrapped DQN}
\label{subsec::atari}

\paragraph{Environments.} To demonstrate the advantages of the mixture usage of different backup lengths, we begin with a collection of well-known \textit{Atari} games, and plot the evaluation curves of eight common \textit{Atari} games selected from the four different categories~\citep{bellemare2016exploration}, including \textit{Breakout}, \textit{Frostbite}, \textit{Hero}, \textit{Enduro}, \textit{Qbert}, \textit{Seaquest}, \textit{CrazyClimber}, and \textit{Freeway}.

\paragraph{Baselines.}
We compare the proposed MB-DQN against two baselines: bootstrapped DQN with (a) all 1-step return heads and (b) all 3-step return heads, which are denoted as \textit{All-1-Step (Baseline)} and \textit{All-3-Step (Baseline)}, respectively. We use ten bootstrapped heads for both MB-DQN and the baselines. MB-DQN (denoted as \textit{Mixed-1-3-Step~(MB-DQN)}) is implemented using five bootstrapped heads with 1-step backup length and another five bootstrapped heads with 3-step backup lengths.

\paragraph{Quantitative comparison.}
The qualitative comparison is presented in Fig.~\ref{fig:atari}. It is observed that \textit{longer backup lengths do not always guarantee better performances --- each environment may have its own favor}. The \textit{All-3-Step} baseline outperforms the \textit{All-1-Step} baseline in six out of eight games, while it underperforms the \textit{All-1-Step} baseline in the rest two games and suffers from a considerable performance drop in \textit{Breakout}. In contrast, the proposed MB-DQN that uses a mixture of step returns outperforms the baselines in terms of its performance and convergence speed for most of the cases.


\begin{figure}[h]
    \centering
    \includegraphics[width=.9\linewidth]{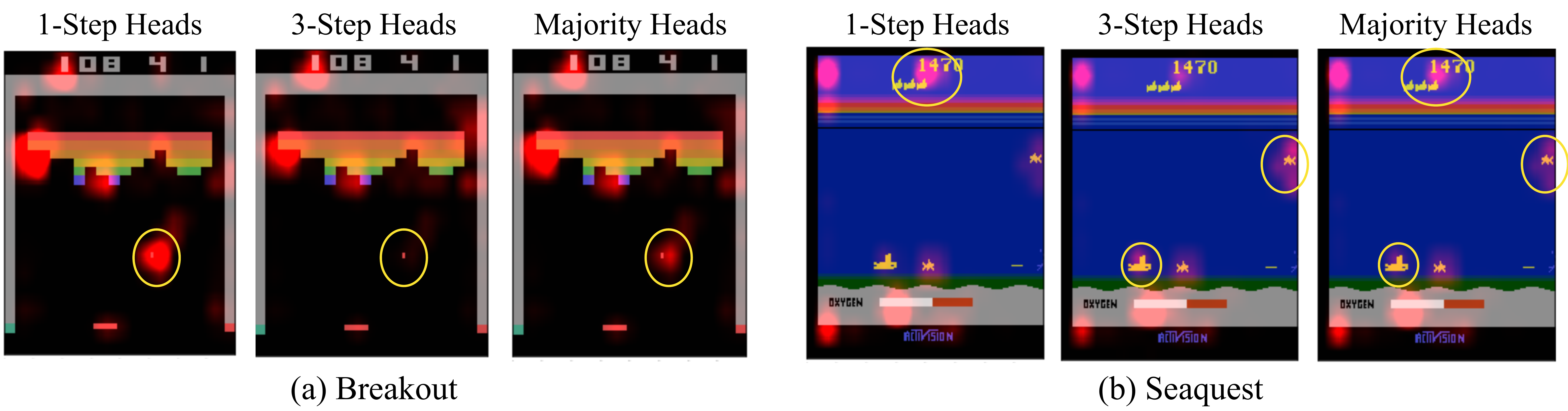}
    \caption{Visualization of the agents’ attention areas (rendered in red) for two \textit{Atari} games. The attention areas are derived based on~\citep{VisaulizeAtari} and are obtained from the bootstrapped heads of MB-DQN.
    }
    \label{fig:viz_atari}
\end{figure}

\paragraph{Qualitative comparison via attention maps.}
To understand the rationale behind the high performance and advantages offered by MB-DQN, we further visualize the attention areas~\citep{VisaulizeAtari} of the agents trained with MB-DQN for three cases: attention areas generated from (a) all of the 1-step bootstrapped heads in MB-DQN (denoted as \textit{1-Step Heads}), (b) all of the 3-step bootstrapped heads in MB-DQN (denoted as \textit{3-Step Heads}), and (c) the composition of the bootstrapped heads which contribute to the decided actions (i.e., the majority of the bootstrapped heads that vote the resultant actions, denoted as \textit{Majority Heads}). Please note that the composition in (c) may contain both the 1-step and 3-step bootstrapped heads. 
Fig.~\ref{fig:viz_atari} illustrates the attention areas (rendered in red) of these three cases for two \textit{Atari} games: \textit{Breakout} and \textit{Seaquest}, and highlights their differences by yellow circles. In \textit{Breakout}, it is observed that the \textit{1-Step Heads} focuses more on the ball, the most important object in this game, than the \textit{3-Step Heads}. It is also observed that when the majority of the heads is considered, the attention of the agent is fell on the ball as well, allowing MB-DQN to play as good as the \textit{All-1-Step} baseline in \textit{Breakout} in Fig.~\ref{fig:atari}.  In \textit{Seaquest}, it is observed that the \textit{1-Step Heads} focuses more on the scoreboard, while the \textit{3-Step Heads} focuses more on the enemy and the submarine. The attention areas of the \textit{Majority Heads}, on the other hand, cover the areas from both the \textit{1-Step Heads} and the \textit{3-Step Heads}, allowing MB-DQN to outperform the two baselines in Fig.~\ref{fig:atari}. These examples thus validate that MB-DQN can leverage the advantages from different backup lengths, and achieve superior performance to the two baselines by offering heterogeneity among its bootstrapped heads.

\subsection{Advantages offered by MB-DQN in the Quality of the Collected Data Samples}
\label{subsec::data_sample_quality}


As the experimental results presented in the previous section have quantitatively and qualitatively demonstrated the performance benefits offered by MB-DQN, we next dive further to investigate the rationale behind the advantages. We hypothesize that the performance improvements provided by MB-DQN may come from the quality of the collected data samples in the experience replay buffer. In other words, MB-DQN may have benefited from the heterogeneity in the data samples collected by bootstrapped heads with different backup lengths. To validate this hypothesis, we design an experiment containing two agents: one agent is responsible for generating state-action pairs for an experience replay buffer while updating its Q-value network with the data contained in it. The other agent only updates its Q-value network by the existing data samples contained in the replay buffer, without contributing data to it. Both of these agents are implemented with ten bootstrapped heads.

\begin{figure}[t]
    \centering
    \includegraphics[width=.9\linewidth]{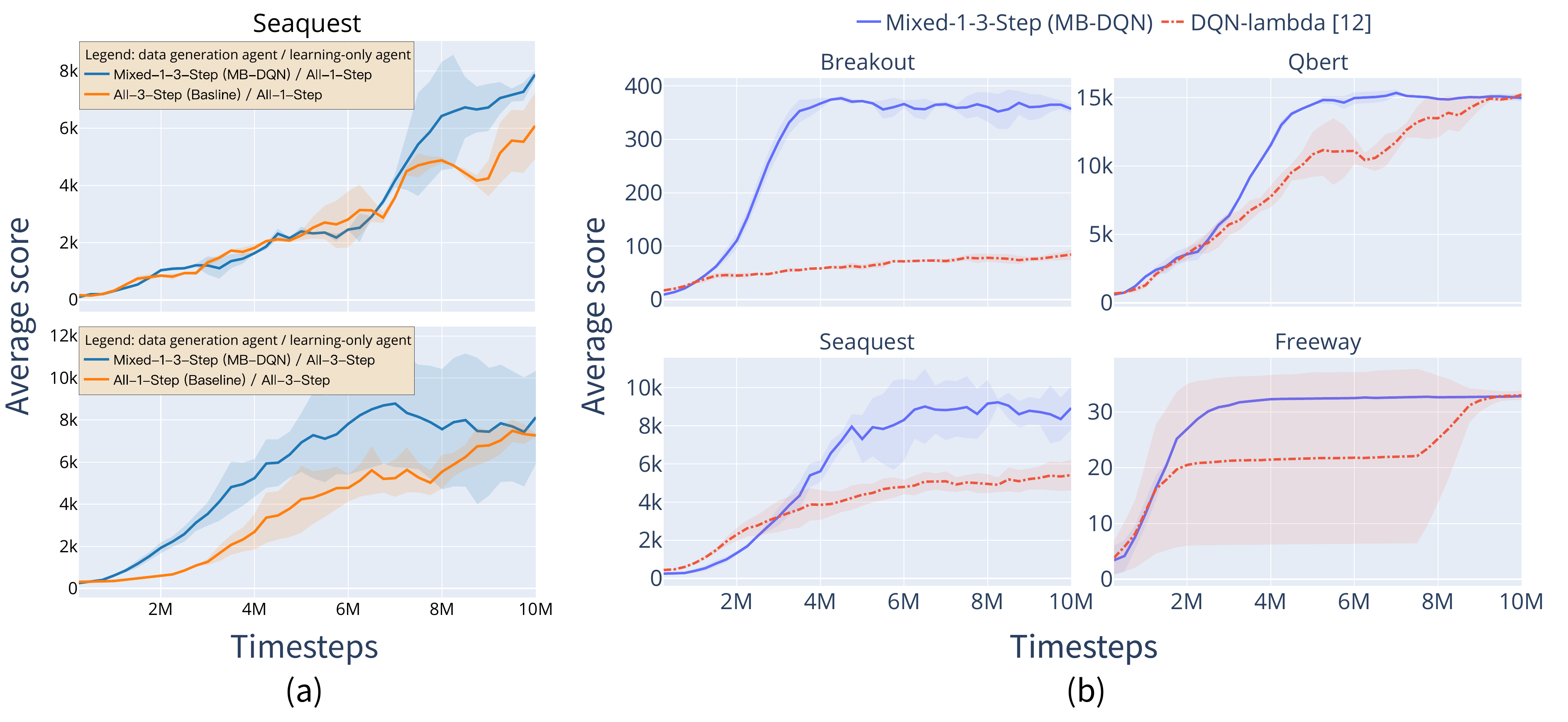}
    \caption{The evaluation curves for two experiments: (a) comparison of different configurations of \textit{data generation agents} and \textit{learning-only agents} for validating the quality of data samples collected by MB-DQN in Section~\ref{subsec::data_sample_quality}, and (b) comparison between the single $\lambda$-target strategy adopted by DQN~($\lambda$)~\citep{DQNlambda} and the multiple bootstrapped targets strategy adopted by MB-DQN in Section~\ref{subsec::target_comparison}.}
    \label{fig:megumi}
\end{figure}

We consider three configurations for the former \textit{data generation agent}: \textit{Mixed-1-3-Step~(MB-DQN)}, \textit{All-1-Step (Baseline)}, and \textit{All-3-Step (Baseline)}, and two configurations for the latter \textit{learning-only agent} (i.e., the one without contributing data samples to the replay buffer): \textit{All-1-Step (Baseline)} and \textit{All-3-Step (Baseline)}. 
The configurations are evaluated on \textit{Seaquest}, and are designed such that the former and the latter agents have different configurations. The results of our experiments are plotted in Fig.~\ref{fig:megumi}~(a), where the figures on the top and bottom sides correspond to the cases that the \textit{learning-only agents} are configured to \textit{All-1-Step (Baseline)} and \textit{All-3-Step (Baseline)}, respectively. It can be observed that the \textit{learning-only agents} trained with the data samples generated by \textit{Mixed-1-3-Step~(MB-DQN)} outperform the agents trained with the data samples generated by the other configurations in both cases. These results thus validate our hypothesis that the data samples generated by MB-DQN are superior in quality than those generated by the other configurations, and explain why MB-DQN is able to offer benefits in performance in the environments presented in this paper.

\subsection{Single $\lambda$-Target versus Multiple Bootstrapped Targets}
\label{subsec::target_comparison}

In order to validate our assumption in Section~\ref{sec::introduction} that unifying different step return targets to a single target value may not be as effective as the bootstrapped approach adopted by MB-DQN, in this section, we compare these two strategies of combining step returns in several \textit{Atari} environments. For the unified return target strategy, we consider a recently proposed method called DQN~($\lambda$)~\citep{DQNlambda}, which implements TD~($\lambda$) by pre-computing $\lambda$-returns using an additional cache for its replay buffer memory. On the other hand, MB-DQN employs a strategy that leverages $K$ bootstrapped heads, where each head $k\in K$ has its own target value. In our experiments, $K$ is set to ten for MB-DQN, while the settings for DQN~($\lambda$) are configured as its default values, where the single target value used by DQN~($\lambda$) is derived from multiple backup lengths ranging from one to a hundred. The evaluation curves of these strategies are plotted in Fig.~\ref{fig:megumi}~(b). It can be observed that for the four environments presented in Fig.~\ref{fig:megumi}~(b), the curves corresponding to the multiple bootstrapped targets strategy (i.e., MB-DQN) grow faster and higher than those corresponding to the single-unified target strategy (i.e., DQN~($\lambda$)). The above interesting evidence not only validates our assumption in Section~\ref{sec::introduction}, but also reveals that the advantages offered by heterogeneity in multiple target values may outweigh the advantages offered by a single TD~($\lambda$) target that aggregates returns from the long temporal horizon.


\begin{figure}[h]
    \centering
    \includegraphics[width=\linewidth]{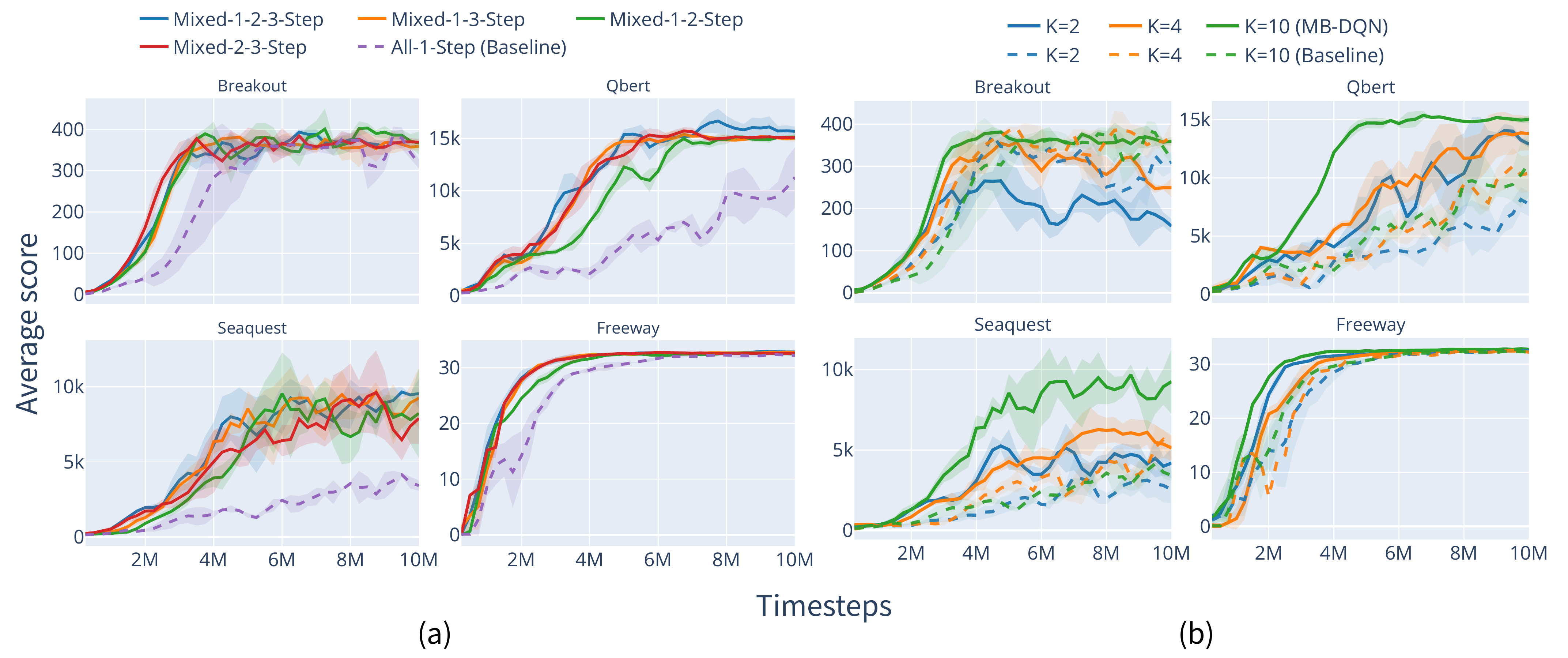}
    \caption{Impacts of (a) different configurations of step returns for the bootstrapped heads, and (b) different numbers of the bootstrapped heads on the proposed MB-DQN for four different \textit{Atari} games.}
    \label{fig:ablation_study}
\end{figure}

\subsection{Ablation Analysis}
\label{subsec::ablation}
In this section, we provide a set of ablation analyses for the proposed MB-DQN on four selceted \textit{Atari} games, including \textit{Breakout}, \textit{Qbert}, \textit{Seaquest}, and \textit{Freeway}, to examine the impacts of configurations on MB-DQN's performances. We perform two sets of analyses for MB-DQN: (a) different configurations of step returns for the bootstrapped heads, and then (b) different numbers of the bootstrapped heads. 


\subsubsection{Different Configurations of Step Returns for the Bootstrapped Heads in MB-DQN}
\label{subsubsec::config_step_returns}

We consider various step returns less than or equal to three, and analyze four different configurations of step returns for the bootstrapped heads in MB-DQN, including \textit{Mixed-1-2-3-Step}, \textit{Mixed-1-2-Step}, \textit{Mixed-1-3-Step}, \textit{Mixed-2-3-Step}, and \textit{All-1-Step (Baseline)}. All of these configurations contain ten bootstrapped heads. The configuration \textit{Mixed-1-2-3-Step} consists of three 1-step bootstrapped heads, three 2-step bootstrapped heads, and four 3-step bootstrapped heads. For the configurations \textit{Mixed-1-2-Step}, \textit{Mixed-1-3-Step}, and \textit{Mixed-2-3-Step}, the ten bootstrapped heads are evenly distributed to different backup lengths.  The results are presented in Fig.~\ref{fig:ablation_study} (a). For all of the configurations, it is observed that the agents trained with different mixtures of step returns perform similarly, and outperform those trained with the \textit{All-1-Step} baseline. These evaluation results thus suggest that the bootstrapped heads in the proposed MB-DQN is not limited to certain configurations of step returns.



\subsubsection{Different Numbers of the Boostrapped Heads} 
In bootstrapped DQN~\citep{BstrapDQN}, more bootstrapped heads lead to faster learning, while even a small number of bootstrapped heads can still capture most of its benefits. As MB-DQN inherits its architecture from bootstrapped DQN, we investigate the impacts of different numbers of the bootstrapped heads on MB-DQN, and examine if MB-DQN still maintains this property or not. We perform experiments with three different numbers of the bootstrapped heads $K = 2, 4,$ and $ 10$. For each configuration, MB-DQN consists of $K/2$ 1-step bootstrapped heads and $K/2$ 3-step bootstrapped heads. The bootstrapped DQN baseline is set as default and is implemented with $K$ 1-step bootstrapped heads.


Fig.~\ref{fig:ablation_study}~(b) illustrates the evaluation curves of the above configurations. In most cases, it is observed that for both MB-DQN and bootstrapped DQN, more bootstrapped heads lead to better performance. It is worth noticing that the proposed MB-DQN trained with two bootstrapped heads outperforms the baseline trained with ten bootstrapped heads in three of four games. This fact shows the significance and advantage of the mixture usage of multi-step returns in bootstrapped heads. On the other hand, MB-DQN's performance drops in \textit{Breakout} as the number of the bootstrapped heads $K$ become less than that of the baseline. This is caused by the fact that the \textit{3-Step Heads} perform worse than the \textit{1-Step Heads} in \textit{Breakout}, as described in Section~\ref{subsec::atari}. As a result, a smaller $K$ strengthens the negative influence caused by the \textit{3-Step Heads}, causing MB-DQN to become sensitive to the undesirable performance in certain bootstrapped heads.  The above results thus suggest that an appropriate number of $K$ has to be selected in order to maintain the advantages as well as the performance of MB-DQN.

\section{Conclusion}
\label{sec::conclusion}

In this paper, we proposed MB-DQN for combining and leveraging the advantages of different step return targets using multiple bootstrapped heads.  Instead of unifying different step return targets to a single target value, MB-DQN assigns a distinct backup length to each bootstrapped head. This allows MB-DQN to offer heterogeneity in the target values during its update procedure, and enables a DRL agent to have diversified exploration behaviors. In our experiments, we first provided motivational examples to demonstrate the influence of different configurations of backup lengths in a simple maze environment. We then evaluated the proposed MB-DQN methodology on a number of \textit{Atari 2600} environments both quantitatively and qualitatively, and validated that MB-DQN is able to outperform a number of baseline methods with different configurations of backup lengths.  Finally, we presented a set of ablation studies to inspect the impacts of different design configurations of MB-DQN in detail.


\begin{thebibliography}{16}
\providecommand{\natexlab}[1]{#1}
\providecommand{\url}[1]{\texttt{#1}}
\expandafter\ifx\csname urlstyle\endcsname\relax
  \providecommand{\doi}[1]{doi: #1}\else
  \providecommand{\doi}{doi: \begingroup \urlstyle{rm}\Url}\fi

\bibitem[Mnih et~al.(2015)Mnih, Kavukcuoglu, Silver, Rusu, Veness, Bellemare,
  Graves, Riedmiller, Fidjeland, Ostrovski, Petersen, Beattie, Sadik,
  Antonoglou, King, Kumaran, Wierstra, Legg, and Hassabis]{dqn}
Volodymyr Mnih, Koray Kavukcuoglu, David Silver, Andrei~A. Rusu, Joel Veness,
  Marc~G. Bellemare, Alex Graves, Martin Riedmiller, Andreas~K. Fidjeland,
  Georg Ostrovski, Stig Petersen, Charles Beattie, Amir Sadik, Ioannis
  Antonoglou, Helen King, Dharshan Kumaran, Daan Wierstra, Shane Legg, and
  Demis Hassabis.
\newblock Human-level control through deep reinforcement learning.
\newblock \emph{Nature}, pages 529--533, 2015.

\bibitem[Wang et~al.(2016)Wang, Schaul, Hessel, Hasselt, Lanctot, and
  Freitas]{duelingDQN}
Ziyu Wang, Tom Schaul, Matteo Hessel, Hado Hasselt, Marc Lanctot, and Nando
  Freitas.
\newblock Dueling network architectures for deep reinforcement learning.
\newblock In \emph{Proceedings of International Conference on Machine Learning
  (ICML)}, pages 1995--2003, 2016.

\bibitem[Hasselt et~al.(2016)Hasselt, Guez, and Silver]{doubleDQN}
Hado~van Hasselt, Arthur Guez, and David Silver.
\newblock Deep reinforcement learning with double q-learning.
\newblock In \emph{Proceedings of AAAI Conference on Artificial Intelligence
  (AAAI)}, page 2094–2100, 2016.

\bibitem[Osband et~al.(2016)Osband, Blundell, Pritzel, and Van~Roy]{BstrapDQN}
Ian Osband, Charles Blundell, Alexander Pritzel, and Benjamin Van~Roy.
\newblock Deep exploration via bootstrapped dqn.
\newblock In \emph{Proceedings of Advances in Neural Information Processing
  Systems (NIPS)}, pages 4026--4034. 2016.

\bibitem[Hessel et~al.(2018)Hessel, Modayil, van Hasselt, Schaul, Ostrovski,
  Dabney, Horgan, Piot, Azar, and Silver]{rainbowDQN}
Matteo Hessel, Joseph Modayil, Hado van Hasselt, Tom Schaul, Georg Ostrovski,
  Will Dabney, Dan Horgan, Bilal Piot, Mohammad~Gheshlaghi Azar, and David
  Silver.
\newblock Rainbow: Combining improvements in deep reinforcement learning.
\newblock In \emph{Proceedings of AAAI Conference on Artificial Intelligence
  (AAAI)}, pages 3215--3222, 2018.

\bibitem[Sutton and Barto(1998)]{Sutton1998}
Richard~S. Sutton and Andrew~G. Barto.
\newblock \emph{Reinforcement Learning: An Introduction}.
\newblock The MIT Press, 1998.

\bibitem[Mnih et~al.(2016)Mnih, Badia, Mirza, Graves, Lillicrap, Harley,
  Silver, and Kavukcuoglu]{a3c}
Volodymyr Mnih, Adria~Puigdomenech Badia, Mehdi Mirza, Alex Graves, Timothy
  Lillicrap, Tim Harley, David Silver, and Koray Kavukcuoglu.
\newblock Asynchronous methods for deep reinforcement learning.
\newblock In \emph{Proceedings of International Conference on Machine Learning
  (ICML)}, pages 1928--1937, 2016.

\bibitem[Barth{-}Maron et~al.(2018)Barth{-}Maron, Hoffman, Budden, Dabney,
  Horgan, TB, Muldal, Heess, and Lillicrap]{D4PG}
Gabriel Barth{-}Maron, Matthew~W. Hoffman, David Budden, Will Dabney, Dan
  Horgan, Dhruva TB, Alistair Muldal, Nicolas Heess, and Timothy~P. Lillicrap.
\newblock Distributed distributional deterministic policy gradients.
\newblock In \emph{Proceedings of International Conference on Learning
  Representations (ICLR)}, 2018.

\bibitem[Lillicrap et~al.(2016)Lillicrap, Hunt, Pritzel, Heess, Erez, Tassa,
  Silver, and Wierstra]{DDPG}
Timothy~P. Lillicrap, Jonathan~J. Hunt, Alexander Pritzel, Nicolas Heess, Tom
  Erez, Yuval Tassa, David Silver, and Daan Wierstra.
\newblock Continuous control with deep reinforcement learning.
\newblock In \emph{Proceedings of International Conference on Learning
  Representations (ICLR)}, 2016.

\bibitem[Asis et~al.(2018)Asis, Hernandez{-}Garcia, Holland, and
  Sutton]{unifying}
Kristopher~De Asis, J.~Fernando Hernandez{-}Garcia, G.~Zacharias Holland, and
  Richard~S. Sutton.
\newblock Multi-step reinforcement learning: {A} unifying algorithm.
\newblock In \emph{Proceedings of AAAI Conference on Artificial Intelligence},
  pages 2902--2909, 2018.

\bibitem[Amiranashvili et~al.(2018)Amiranashvili, Dosovitskiy, Koltun, and
  Brox]{TDnotTD}
Artemij Amiranashvili, Alexey Dosovitskiy, Vladlen Koltun, and Thomas Brox.
\newblock Analyzing the role of temporal differencing in deep reinforcement
  learning.
\newblock In \emph{Proceedings of International Conference on Learning
  Representations (ICLR)}, 2018.

\bibitem[Daley and Amato(2019)]{DQNlambda}
Brett Daley and Christopher Amato.
\newblock Reconciling $\lambda$ -- returns with experience replay.
\newblock In \emph{Proceedings of Advances in Neural Information Processing
  Systems (NIPS)}, pages 1133--1142. 2019.

\bibitem[Bellemare et~al.(2015)Bellemare, Naddaf, Veness, and Bowling]{Atari}
Marc~G. Bellemare, Yavar Naddaf, Joel Veness, and Michael Bowling.
\newblock The arcade learning environment: An evaluation platform for general
  agents.
\newblock In \emph{Proceedings of International Joint Conference on Artificial
  Intelligence (IJCAI)}, pages 4148--4152, 2015.

\bibitem[Jaakkola et~al.(1994)Jaakkola, Jordan, and Singh]{VarianceTradeoff}
Tommi~S. Jaakkola, Michael~I. Jordan, and Satinder~P. Singh.
\newblock On the convergence of stochastic iterative dynamic programming
  algorithms.
\newblock \emph{Neural Computation}, 6\penalty0 (6):\penalty0 1185--1201, 1994.

\bibitem[Bellemare et~al.(2016)Bellemare, Srinivasan, Ostrovski, Schaul,
  Saxton, and Munos]{bellemare2016exploration}
Marc~G. Bellemare, Sriram Srinivasan, Georg Ostrovski, Tom Schaul, David
  Saxton, and Rémi Munos.
\newblock Unifying count-based exploration and intrinsic motivation.
\newblock In \emph{Proceedings of Advances in Neural Information Processing
  Systems (NIPS)}, pages 1471--1479, Dec. 2016.

\bibitem[Greydanus et~al.(2018)Greydanus, Koul, Dodge, and
  Fern]{VisaulizeAtari}
Samuel Greydanus, Anurag Koul, Jonathan Dodge, and Alan Fern.
\newblock Visualizing and understanding atari agents.
\newblock In \emph{Proceedings of International Conference on Machine Learning
  (ICML)}, volume~80, pages 1787--1796, 2018.

\end{thebibliography}
\end{document}